\documentclass[final,3p,times]{elsarticle}

\DeclareGraphicsExtensions{.jpg,.pdf,.mps,.png} 
\graphicspath{{img/} {./}} 

\usepackage{amssymb}
\usepackage{amsthm}

\usepackage{lineno}
\usepackage{url,hyperref,lineno,microtype,subcaption}
\usepackage[onehalfspacing]{setspace}
\usepackage{enumerate}
\usepackage{caption}
\usepackage{subcaption}
\usepackage{multirow}

\begin{document}

\begin{frontmatter}



\title{A lunar reconnaissance drone for cooperative exploration and high-resolution mapping of extreme locations}


\author[espace]{Roméo Tonasso}
\author[espace]{Daniel Tataru} 
\author[espace]{Hippolyte Rauch}
\author[espace]{Vincent Pozsgay}
\author[espace]{Thomas Pfeiffer}
\author[espace]{Erik Uythoven}
\author[aqua]{David Rodríguez-Martínez}

\affiliation[espace]{organization={eSpace - EPFL Space Center, École Polytechnique Fédérale de Lausanne (EPFL)},
            city={Lausanne},
            postcode={1015}, 
            country={Switzerland}}
            
\affiliation[aqua]{organization={Advanced Quantum Architecture Laboratory (AQUA), École Polytechnique Fédérale de Lausanne (EPFL)},
            city={Neuchâtel},
            postcode={2000}, 
            country={Switzerland}}

\begin{abstract}
An efficient characterization of scientifically significant locations is essential prior to the return of humans to the Moon. The highest resolution imagery acquired from orbit of south-polar shadowed regions and other relevant locations remains, at best, an order of magnitude larger than the characteristic length of most of the robotic systems to be deployed. This hinders the planning and successful implementation of prospecting missions and poses a high risk for the traverse of robots and humans, diminishing the potential overall scientific and commercial return of any mission. We herein present the design of a lightweight, compact, autonomous, and reusable lunar reconnaissance drone capable of assisting other ground-based robotic assets, and eventually humans, in the characterization and high-resolution mapping ($\sim$\,0.1\;m/px) of particularly challenging and hard-to-access locations on the lunar surface. The proposed concept consists of two main subsystems: the drone and its service station. With a total combined wet mass of 100\;kg, the system is capable of 11 flights without refueling the service station, enabling almost 9\;km of accumulated flight distance. The deployment of such a system could significantly impact the efficiency of upcoming exploration missions, increasing the distance covered per day of exploration and significantly reducing the need for recurrent contacts with ground stations on Earth.
\end{abstract}




\end{frontmatter}


\section{Introduction}
\label{sec:intro}

NASA has recently selected 13 candidate landing sites in the south polar region of the Moon for their Artemis III mission \citep{nasa2022}, a mission aimed at sending the first group of humans to the lunar surface since the Apollo program. Prior to human exploration of the Moon and in line with the goals of the new Artemis program, a series of upcoming robotic missions spearheaded by both national space agencies and private corporations are also aiming at characterizing and prospecting a number of relevant locations on the lunar surface. Among these, south-polar Permanently and Transiently Shadowed Regions (PSRs and TSRs, respectively) and lunar skylights appear as primary candidates, potentially bearing answers to fundamental questions on the origin and formation of the Moon \citep{wagner2014, michael2018}, harboring valuable resources for in-situ extraction \citep{crawford2015, cannon2020, brown2022, cannon2023}, and providing shelter beyond Earth where humans could finally settle \citep{blair2017}. To accomplish all of the above, \textbf{efficient exploration} of scientifically and commercially significant locations is essential.

Efficient exploration means deploying highly autonomous robotic systems with the capacity to traverse longer distances (\textgreater\,100\;km) under increasingly constrained time windows (e.g., shorter day-light cycles on high-latitude regions due to the low lunar obliquity), to effectively operate under extreme environments (i.e., across unstructured, dynamic, and hazard-abundant landscapes with, at times, lack of natural illumination, cryogenic temperatures, and subject to the impact of meteorites and high-energy radiation) of which fewer and/or lower quality data are readily available, and to do so in a cost-effective manner (e.g., \textgreater\,6\;meters/\$100k invested for a single lunar mission). 

One of the key enablers of efficient exploration is having access to high-resolution topographical and geomorphological data. The highest resolution images of the lunar surface acquired from lunar orbit to date have been measured by the Narrow Angle Cameras (NACs) onboard NASA's Lunar Reconnaissance Orbiter (LRO). NACs are capable of mapping regions of the Moon down to a spatial resolution of 0.5\;m/px \citep{robinson2010}. This is achieved, however, under optimal lighting conditions. When resolving internal features of PSRs and TSRs, the prospect of achieving this level of resolution from orbit is unlikely. Imaging shadowed and poorly-lit areas on the surface requires longer exposure times, which paired with the increased shot noise and rapid movement of the satellites drastically worsens the overall signal-to-noise ratio (SNR) of the output images \citep{cisneros2017, martin2021}. Images taken by the LRO of unlit regions on the Moon display maximum spatial resolutions after resampling of $\sim$\,10\;m/px \citep{martin2020}. Similar results were previously achieved by the Terrain Camera (TC) onboard JAXA's ``Kaguya'' Selenological and Engineering Explorer (SELENE) \citep{haruyama2008}. More recently, NASA's ShadowCam instrument currently operating onboard KARI's Korea Pathfinder Lunar Orbiter (KPLO) was specifically developed to capture images of PSRs at a maximum spatial resolution of 1.7\;m/px \citep{shadowcam2021}. And new learning-based image post-processing approaches, such as the Hyper-effect nOise Removal U-net Software (HORUS) \citep{bickel2021} developed by a team from ETH Zurich, University of Oxford, and NASA Ames Research Center, are being devised to artificially enhance the SNR of existing data sets while achieving improved spatial resolutions ($\sim$\,1\;m/px) on long-exposure images \citep{bickel2022}. 

\begin{figure}[h!]
\begin{center}
\includegraphics[width=0.8\textwidth]{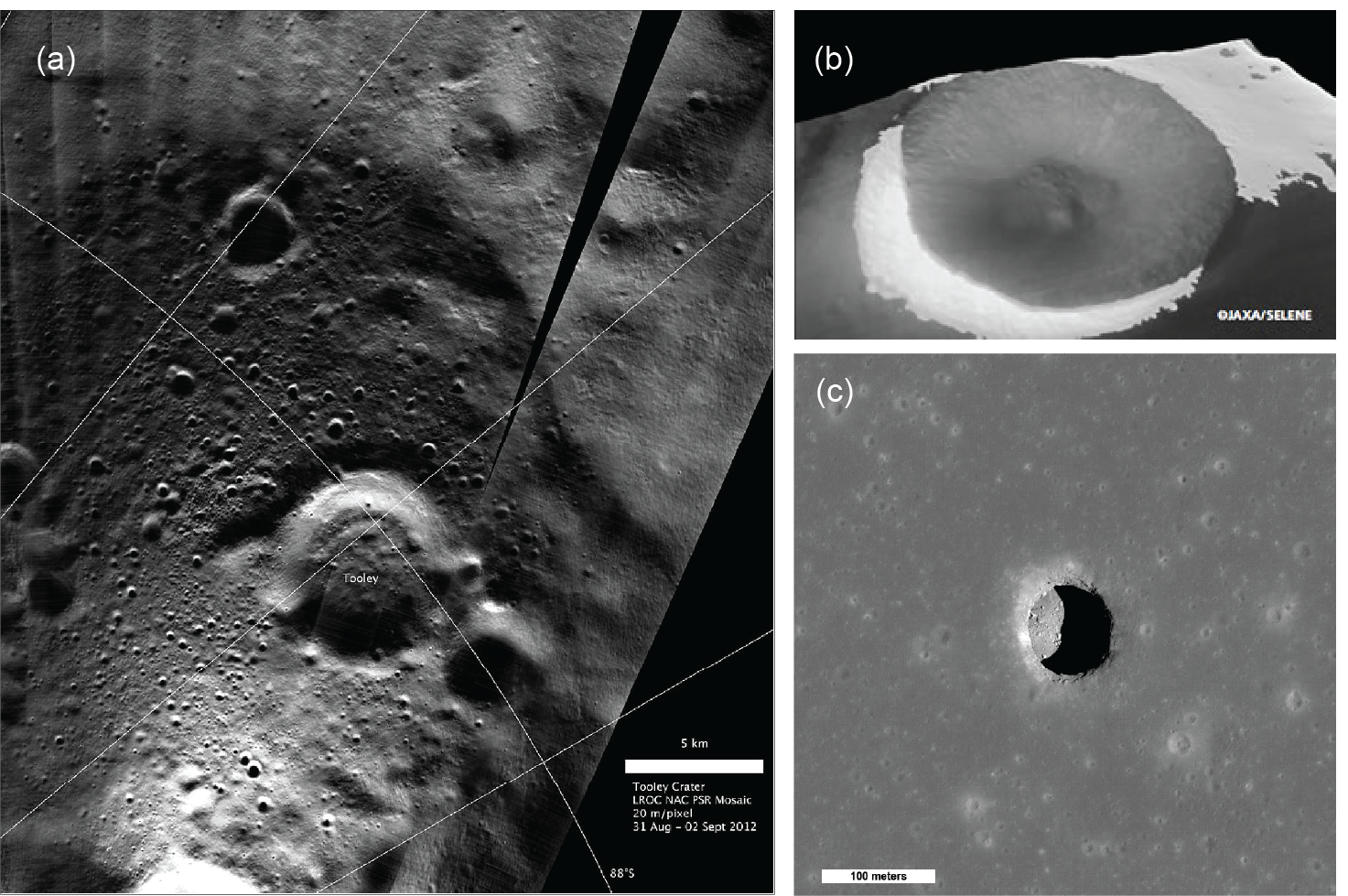}
\end{center}
\caption{Examples of some of the highest resolution images available of PSRs and lunar skylights: (a) a long exposure of the permanently shadowed Tooley Crater taken by LRO NAC (source: NASA/Goddard/Arizona State University), (b) an enhanced image of the Shackleton Crater taken by Kaguya's TC (source: JAXA), and (c) view of a skylight at Mare Tranquillitatis also taken by LRO NAC (source: NASA/Goddard/Arizona State University).}\label{fig:orbital_imgs}
\end{figure}

Even though new technologies and approaches are significantly improving the quality of orbital measurements, current maps of these highly relevant lunar regions are still too coarse for an optimal and efficient mission planning. Data at spatial resolutions equivalent to that of a factor of the characteristic length of the systems to be deployed---i.e., wheelbase, wheel track, or even wheel size for wheeled robots and stride or step length for legged robots and potentially humans---are required. The impossibility to resolve sub-meter hazards and/or precisely pinpoint local regions of interest from these images negatively impacts the efficacy and effectiveness of these missions, precluding the possibility to cover large distances, increasing overall mission risk, and diminishing the potential scientific or commercial return on investment on any given mission. 

We present an alternative to traditional single-rover missions and previously presented concepts for long-distance coverage (see Section~\ref{sec:background} for details). Our concept aims at solving the issue of high-resolution data acquisition at large scales. In the following pages, we describe the outcome of a feasibility analysis and preliminary design study on the potential deployment of a lunar reconnaissance drone for exploring, characterizing, and high-resolution mapping ($\sim$\,0.1\;m/px) of targeted regions of interest.

\section{Background}
\label{sec:background}
The miniaturization of electromechanical components has rapidly impacted the development of small-sized, lightweight unmanned aerial vehicles (UAVs) on Earth. The spectrum of applications for which terrestrial drones are being used is constantly widening: from emergency management and surveillance \citep{erdelj2017} to marine monitoring \citep{yang2022}. UAVs benefit from ease of operation, fast deployment, and long-distance coverage while being economical and transportable. 

Beyond Earth, the deployment of UAVs, or aerobots as they are often referred to in planetary exploration---a term that includes rotorcraft \citep{zhao2019, pipenberg2022, tzanetos2022a, patel2023}, fixed-wing drones \citep{landis2003, braun2006}, lighter-than-air vehicles \citep{jeffery2007, blamont2008}, and suborbital hoppers \citep{burdick2003, montminy2008, howe2011}---, has been a topic of discussion and conceptualization for exploring atmosphere-bearing celestial bodies ever since the first martian airplane concept was sketched at NASA's Jet Propulsion Laboratory \citep{clarke1979}. Mars Helicopter Ingenuity, part of NASA's Mars 2020 mission \citep{tzanetos2022b}, has recently become the first unpiloted aircraft to perform a power-controlled flight on another planet \citep{crane2021}. Ingenuity's feat has brought about a renewed interest in the use of rotorcraft for exploration, enabling opportunities for new science, and redrawing concepts for upcoming missions to the red planet \citep{muirhead2020, nasa2023}.

On the Moon, however, aerobots demand an extra layer of complexity. The negligible atmosphere present on the Moon \citep{stern1999} requires the use of either electromechanical devices for short-distance skipping and pronking \citep{philip2019} or rocket engines for long-distance hopping and flying. In the latter category, a much lower number of concepts are described in the literature compared to that of martian aircraft. 

\begin{figure}[h!]
\begin{center}
\includegraphics[width=\textwidth]{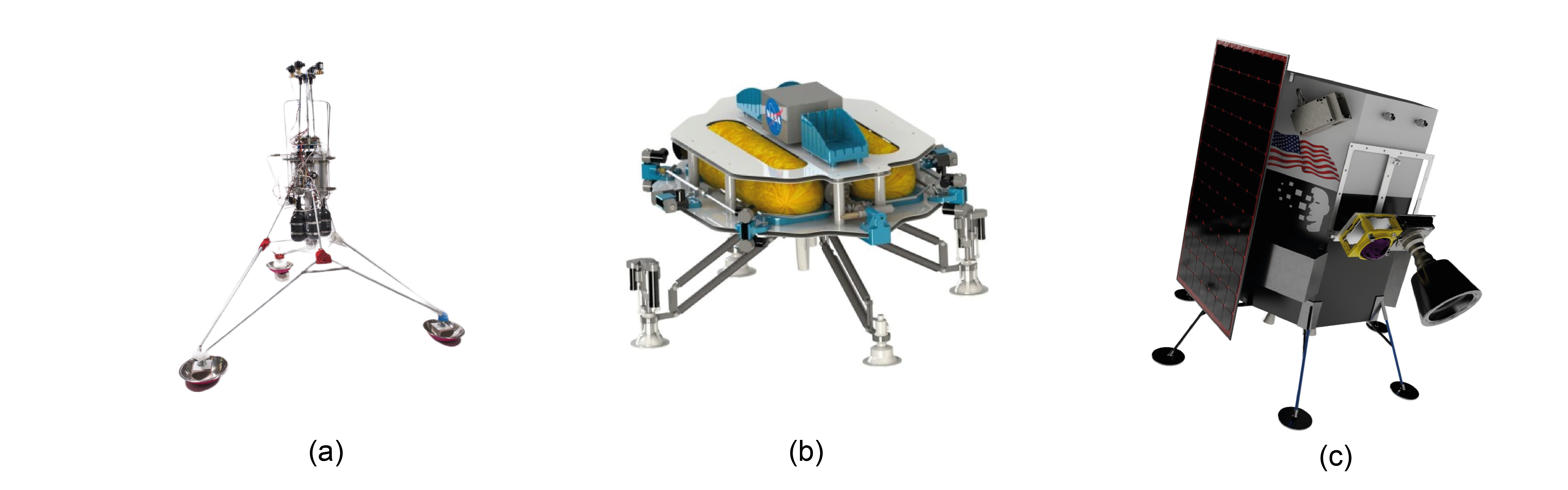}
\end{center}
\caption{Examples of conceptual designs for drones and long-distance hoppers for the exploration of the Moon and other airless celestial bodies: (a) the Lunar Hopper Mk. II (source: University of Southampton), (b) the Extreme Access Flying drone concept (source: NASA/Swamp Works), and (c) the $\mathrm{\mu}$Nova lunar hopper (source: Intuitive Machines).}\label{fig:drones}
\end{figure}

Two studies conducted at the Massachusetts Institute of Technology (MIT) outlined a series of potential mission scenarios, operational concepts, and safe landing approaches for planetary hoppers \citep{nothnagel2011} and described the development of TALARIS \citep{cohanim2013}, a lunar hopper prototyped for Earth-based testing propelled by cold-gas thrusters. Another group of students from the University of Southampton designed and tested a prototype of a Vertical Take-Off \& Vertical Landing (VTVL) lunar hopper \citep{daogaru2016}. This hopper, dubbed Lunar Hopper Mk. II (Figure~\ref{fig:drones}(a)), weighs 37\;kg and it is mainly propelled by a single 400-N hybrid rocket engine and controlled by four nitrogen-based cold-gas thrusters. A group of spherical drones, called SphereX, has been proposed by a team from NASA's Goddard Space Flight Center for the cooperative exploration of underground lava tubes, caves, and other extreme locations \citep{thangavelautham2014}. SphereX robots are meant to be capable of rolling, hopping, and flying. With a diameter of 0.3\;m and a total wet mass of just 3\;kg, each SphereX has an anticipated payload carrying capacity of 1\;kg and about 5\;km of flight range on the Moon. Its propulsion system consists of a bi-propellant (RP1-$\mathrm{H_2O_2}$) engine and eight $\mathrm{H_2O_2}$-based attitude control thrusters. While extensive work has been conducted on the mobility and control of these spherical robots, questions remain unanswered as to the manufacturing and potential miniaturization of the propulsion system \citep{kalita2018, kalita2020}. On the subject of lunar drones, Swamp Works, a group formed by engineers at NASA's Kennedy Space Center, also presented their own concept for what they called Extreme Access Flyers (Figure~\ref{fig:drones}(b)). With a width slightly larger than 150\;cm, these drones are equipped with cold-gas thrusters for take-off and landing (TOL) and attitude control \citep{engelking2015}. Another concept has been introduced by Politecnico di Torino for an autonomous 12U suborbital lunar drone \citep{podesta2020}. This drone would have a total estimated wet mass of 12\;kg and be propelled by experimental $\mathrm{H_2O_2}$-based monopropellant engines in a similar configuration to that of the SphereX (1 main, 8 for attitude control). Similar challenges associated with the miniaturization and maturation of the propulsion technology were found.

In the realm of commercial applications, Intuitive Machines, an American company founded in 2013, has recently signed a contract with NASA for the development of its $\mathrm{\mu}$Nova lunar hopper \citep{valentine2021} (Figure~\ref{fig:drones}(c)), a scaled-down version of the company's lander, Nova-C \citep{fox2021}. Once detached from the lander, the 30-to-50-kg $\mathrm{\mu}$Nova is designed to hop across PSRs and into lunar pits. The system reuses the same precision landing and hazard avoidance sensor suite and software used to land Nova-C on the lunar surface \citep{atwell2022}.


\section{Challenges}
\label{sec:challenges}

The basic premise of our concept is founded on the current use, form factor, and operability of terrestrial UAVs while building on top of the work already conducted on lunar aerobots.  We set out to design a fully autonomous, lightweight, compact, modular, adaptable, and reusable lunar drone capable of cooperating with other robotic assets or vehicles operating on the surface of the Moon. This presented the following challenges: 

\begin{enumerate}[1. ]
    \item Achieving \textbf{full autonomy} implied making the most of the limited computational capacity of existing space-qualified processing units while limiting the extent of sensory input required in flight and the complexity of the trajectories to be followed. 
    \item For the drone to be as \textbf{lightweight and compact} as possible, fuel consumption had to be optimized and the amount of power required onboard needed to be heavily limited (e.g., by avoiding complex active thermal regulation systems but still being able to sustain the extreme thermal fluctuations of PSRs/TSRs \citep{keller2016}). 
    \item \textbf{Modularity, adaptability, and reusability} meant being capable of hosting different instruments for different purposes, being capable of operating alongside multiple platforms in a wide array of mission scenarios, and being capable of achieving multiple flights per mission over multiple missions. 
\end{enumerate}

To further constrain our analysis, we grounded our study on particular features of the upcoming NASA's Volatiles Investigating Polar Exploration Rover (VIPER) Mission \citep{colaprete2019} and ESA's European Large Logistics Lander (EL3) \citep{golling2020}. These introduced the high-level preliminary requirements listed in Table~\ref{tab:requirements}. 

 \begin{table}[hbt!]
    \centering
    \begin{tabular}{l p{5cm} p{5.5cm}}
        \hline
        \textbf{ID} & \textbf{Requirement} & \textbf{Note} \\ 
        \hline
        R1 & The drone shall have a flight range of at least 800\;m. & Minimum flight range has been estimated to be at least double that of the minimum traverse within PSRs of the EL3 Lunar Polar Explorer mission concept rover. \\ \hline
        R2 & The drone system shall survive at least 50 hours of continuous darkness in \textit{standby} mode. & Although it should be mostly inteded to fly above shaded areas, the drone shall be capable of coping with the long, fast-moving shadows characteristic of high-latitude regions. \\ \hline
        R3 & The drone system shall be scaled to survive on the Moon for a minimum of 100 days. & Lifetime estimated based on that of NASA's VIPER rover. \\ \hline
        R4 & The drone shall fit within a maximum allowed volume of 4\;$\mathrm{m^3}$. & Estimation based on the available footprint on top of the VIPER rover$^1$.\\ \hline
        R5 & The drone shall map the surface at resolutions $\sim$~0.1\;m/px. & Target resolution estimated as half that of a common small-to-medium rover wheel size. \\ \hline
        \multicolumn{3}{l}{\tiny{$^1$Our initial concept of operations defined a service station assembled on top of the rover to be assisted (see Section~\ref{sec:service_station})}}
    \end{tabular}    
    \caption{Mission concept high-level requirements}
    \label{tab:requirements}
\end{table}


\section{Concept of Operations}
\label{sec:conops}

\begin{figure}[h!]
\begin{center}
\includegraphics[width=0.9\textwidth]{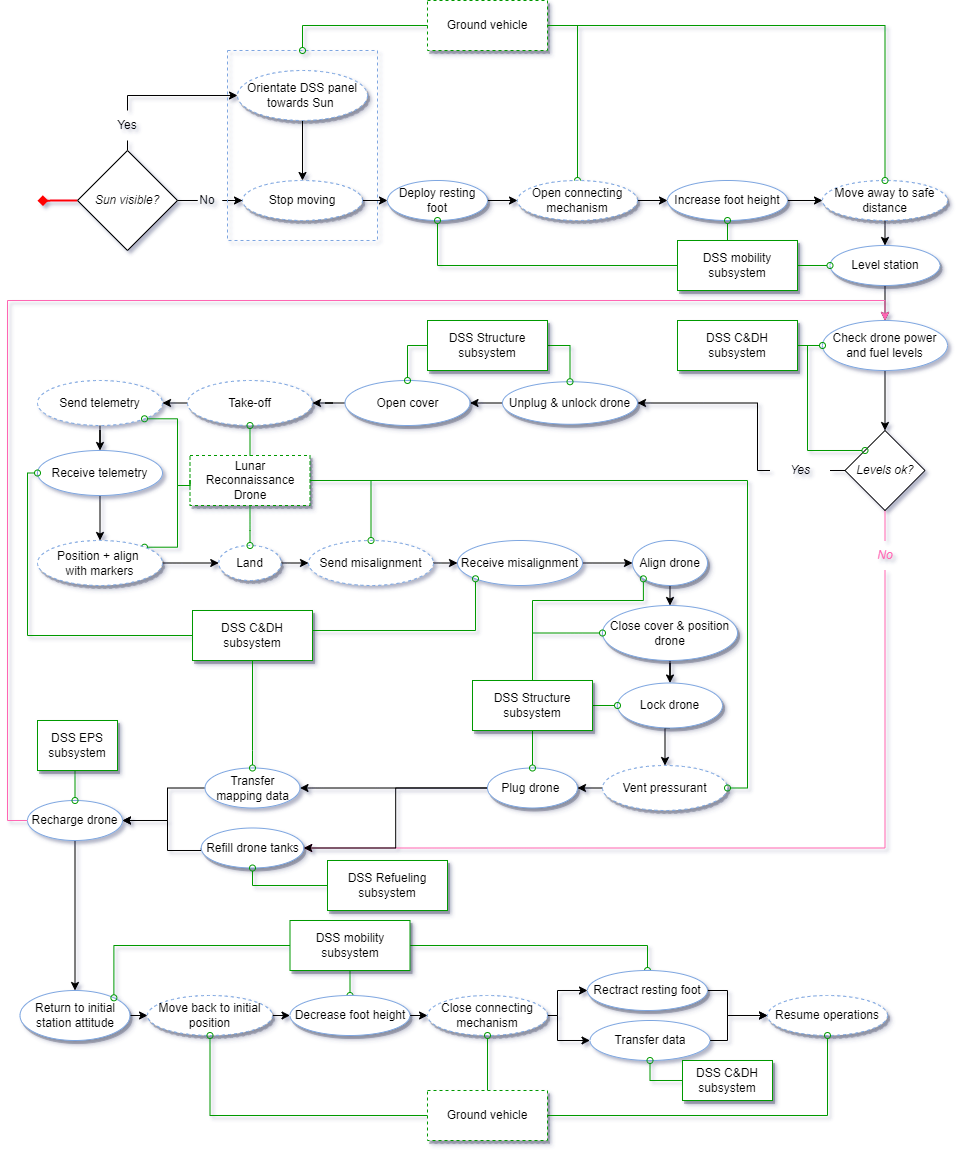}
\end{center}
\caption{Schematic CONOPS for a single reconnaissance flight. Starting point is highlighted in red. Blue ovals indicate actions. Green rectangles indicate actors: the drone, the drone service station (DSS), and the serviced ground vehicle.}
\label{fig:conops_diagram}
\end{figure}

With these challenges in mind, we envisioned a payload envelope (referred to herein as the ``drone system'') formed by the drone and a so-called service station in the form of a towed trailer. In our proposed concept of operations (CONOPS), a prospecting rover approaches a region of which limited geomorphological information is available for an optimal traverse (e.g., a PSR) or one characterized by an extreme topography for the rover to access (e.g., the rim of a crater or the edge of a skylight). The rover detaches from the service station allowing its cover panels to open, revealing and releasing the support structures that hold the drone in place (see Figure~\ref{fig:dss_conops}). The drone is deployed, climbs to an altitude of 50\;m above ground level, and proceeds to follow a predefined trajectory optimized for maximum coverage and minimum fuel consumption, flying to a maximum horizontal distance of 400\;m away from the service station (flight simulations are described in Section~\ref{sec:simulations}). With the data acquired in flight, the drone returns to the original take-off location, landing safely back on the service station. This operation can then be repeated multiple times---up to 11 with our current concept---over the course of any given mission covering local areas where more or higher quality environmental information is needed. Local elevation maps of the surroundings can then be created by the rover or any other ground assets in the surroundings to more effectively characterize the area. A detailed flow chart of these operations is depicted in Figure~\ref{fig:conops_diagram}.

The \textbf{service station} was devised as a necessary multifunctional element of the drone system. Its role is to act as a TOL pad, as a refueling and recharging station for the drone, as a shelter for the drone when not in operation, and as a depot for major data transmissions between the drone and the rover or any other surrounding robots or vehicles. The specifics of the design of the service station are described in Section~\ref{sec:service_station}.

One of the common drawbacks we encountered when evaluating existing concepts and mission architectures (Section~\ref{sec:background}) was the need to always take off, land, or hop from the ground. This has some clear benefits---longer flight range, potentially lower fuel consumption, and/or higher independence. In the case of lunar missions, however, we deemed interacting with the ground a major drawback for the following reasons: 1) its negative impact on potential surface and subsurface volatiles and other valuable elements present within the region of influence of the propulsion system \citep{colaprete2010, paige2010}, 2) having to cope with excessive and slow-settling dust generated by firing the engines close to the ground \citep{canon2022} and its potential effect on orbiting spacecraft \citep{metzger2021}, 3) the need for more sophisticated flight software solutions to enable safe autonomous landing on unknown, unstructured, uneven, and hazardous terrains affected by complex illumination conditions \citep{silvestrini2022}, and 4) the non-negligible impact of extremely low surface temperatures (as low as 20\;K within some PSRs \citep{keller2016}) on the overall size and weight of the system (e.g., the need to implement additional heaters and/or radiators). The concept of the service station came about as a potential solution that mitigates most of these issues. It enables a higher fuel-carrying capacity per mission with the addition of refueling tanks increasing its adaptability to different missions and reducing mission risks by simplifying the avionics since the drone is intended to always TOL from a well-known, flat, and dust-free location. The drone has been designed, however, to be capable of emergency landing on the ground in the event of a failure.

\section{System Design}

\begin{figure}[h!]
\begin{center}
\includegraphics[width=0.8\textwidth]{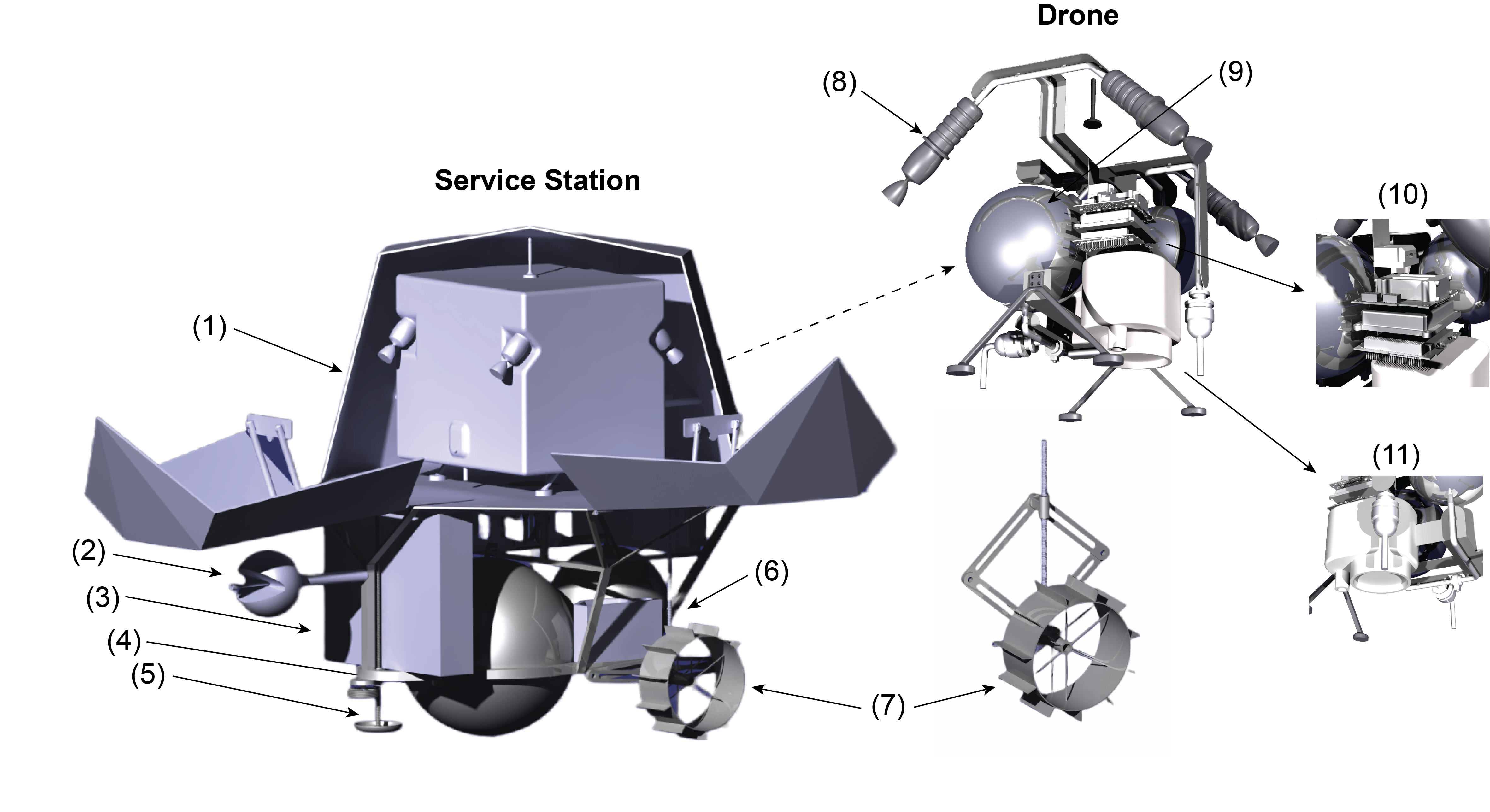}
\end{center}
\caption{The lunar reconnaissance drone system is composed of the drone and its service station. Some key elements of the system include: (1) outer shell (protective panels, TOL pad, interfaces), (2) docking mechanism, (3) batteries, (4) refueling tanks, (5) adjustable resting foot for deployment, (6) electronic box, (7) actuated control arm and wheel, (8) monopropellant thrusters, (9) propellant and pressurant tanks, (10) drone electronic stack (IMU, EPS, OBC, and CDH), and (11) mapping sensor.}
\label{fig:system}
\end{figure}

The system consists of a drone and its service station. The drone system is designed to assist other planetary robots, ground vehicles, and eventually humans operating on the surface into inaccessible environments or those of which scattered, low-resolution data is available. The drone system is designed for fast deployment and ease of operation. It is meant to be a low-cost solution that prevents excessive contamination of pristine locations with high scientific, and potentially high commercial, value. The full system (see Figure~\ref{fig:system}) has an overall wet mass of 100\;kg and in its current configuration provides a total flight range of 9\;km or a total of 11 flights without refueling the station.

\subsection{Lunar Reconnaissance Drone}

\begin{figure}[h!]
\begin{center}
\includegraphics[width=0.8\textwidth]{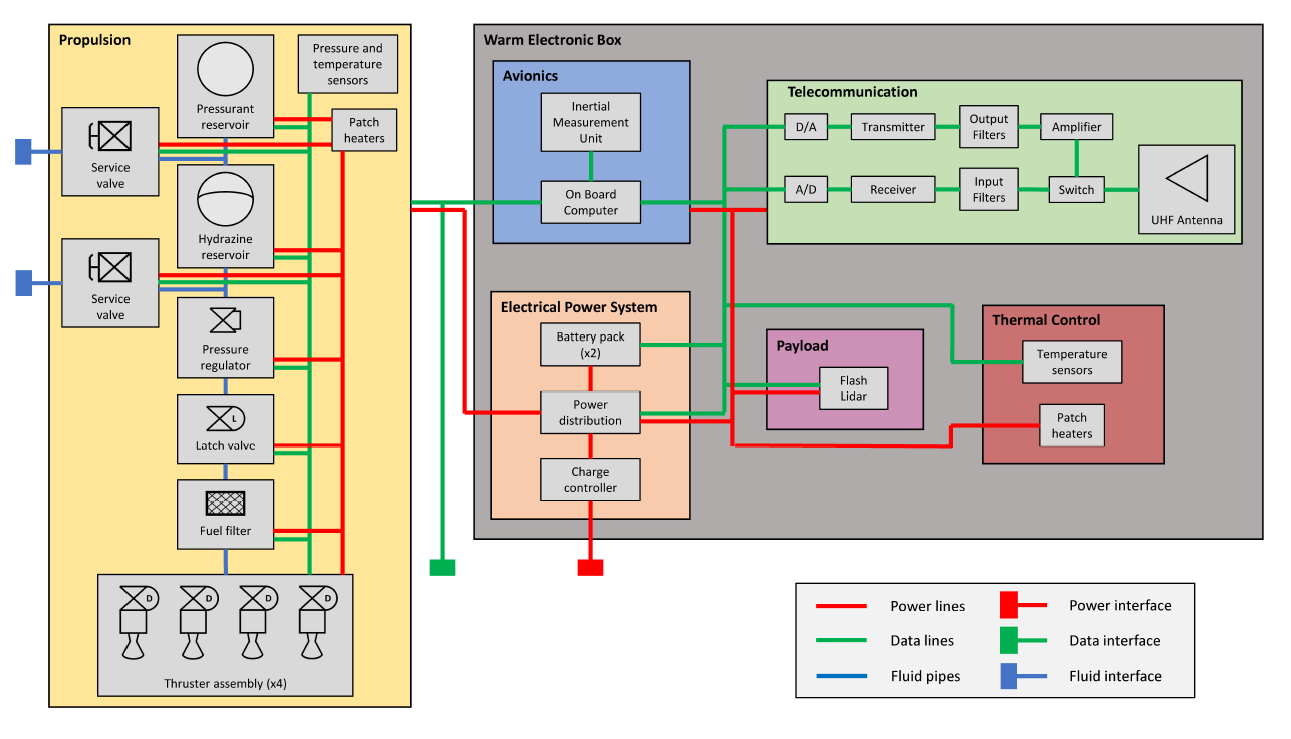}
\end{center}
\caption{Architecture of the lunar reconnaissance drone.}\label{fig:drone_block}
\end{figure}

A high-level schematic of the different subsystems and components comprising the drone is presented in Figure~\ref{fig:drone_block}. Connecting lines illustrate the different internal and external interfaces. The drone has a dimension of 450 x 480 x 378\;mm and a total wet mass of 16.96\;kg, of which 8.15\;kg are devoted to the propulsion system alone, including 2.42\;kg of total propellant and pressurant in a 2.5:1 ratio. The total estimated power consumption of the drone in flight yields 324\;W. A \textit{standby} mode will be used when docked with the service station keeping most of the subsystems either off or in a low-power mode. Details of the design of relevant subsystems are presented in the following sections. Space-qualified off-the-shelf components were favored to define a baseline for the design and size the system whenever possible. 

\subsubsection{Propulsion}

The selection of the propulsion subsystem of the drone is a key driver for the full system design specifications and its operability. The type of propulsion needed had to provide enough thrust while being throttleable in the range between 10--100\;N. In line with the engine technologies favored in previous designs (refer to Section~\ref{sec:background}), we ultimately opted for a system formed by four 22-N MR-106L monopropellant thrusters \citep{aerojet2020} fueled by hydrazine and a S-405 catalyst. The main specifications of these engines are listed in Table~\ref{tab:thrusters}.

Monopropellant engines provide enough thrust, compared to electrical engines, while being refuelable, unlike hybrid engines. They present a good balance between simplicity and low mass compared to that of bi-propellant rocket engines and provide a higher specific impulse (ISP) than cold gas systems. A rapid and precise control of the drone also demanded a low minimum impulse bit (MIB) ($\leq$~80\;mN·s based on preliminary simulations, refer to Section~\ref{sec:simulations}).

 \begin{table}[hbt!]
    \centering
    \begin{tabular}{p{5cm} p{2.5cm} }
        \hline
        \textbf{Specification} & \textbf{Value} \\ 
        \hline
        Propellant & $\mathrm{N_2H_4}$/S-405 \\ 
        Inlet pressure range & 5.9--27.6\;Bar\\ 
        Thrust range & 10--34\;N\\ 
        Minimum impulse bit & 15\;mN·s \\ 
        Nozzle expansion ration & 60:1 \\ 
        Steady-state ISP (in vacuum) & 228--235\;s \\ 
        Overall length & 186\;mm \\ 
        Mass & 0.59\;kg \\ 
        Pull-in voltage & 36$\pm$4 VDC \\ 
        Steady-state firing & 4000\;s \\ 
        TRL & 9\\ \hline
    \end{tabular}    
    \caption{Thrusters' main specifications \citep{aerojet2020}.}
    \label{tab:thrusters}
\end{table}

\begin{figure}[h!]
\begin{center}
\includegraphics[width=0.5\textwidth]{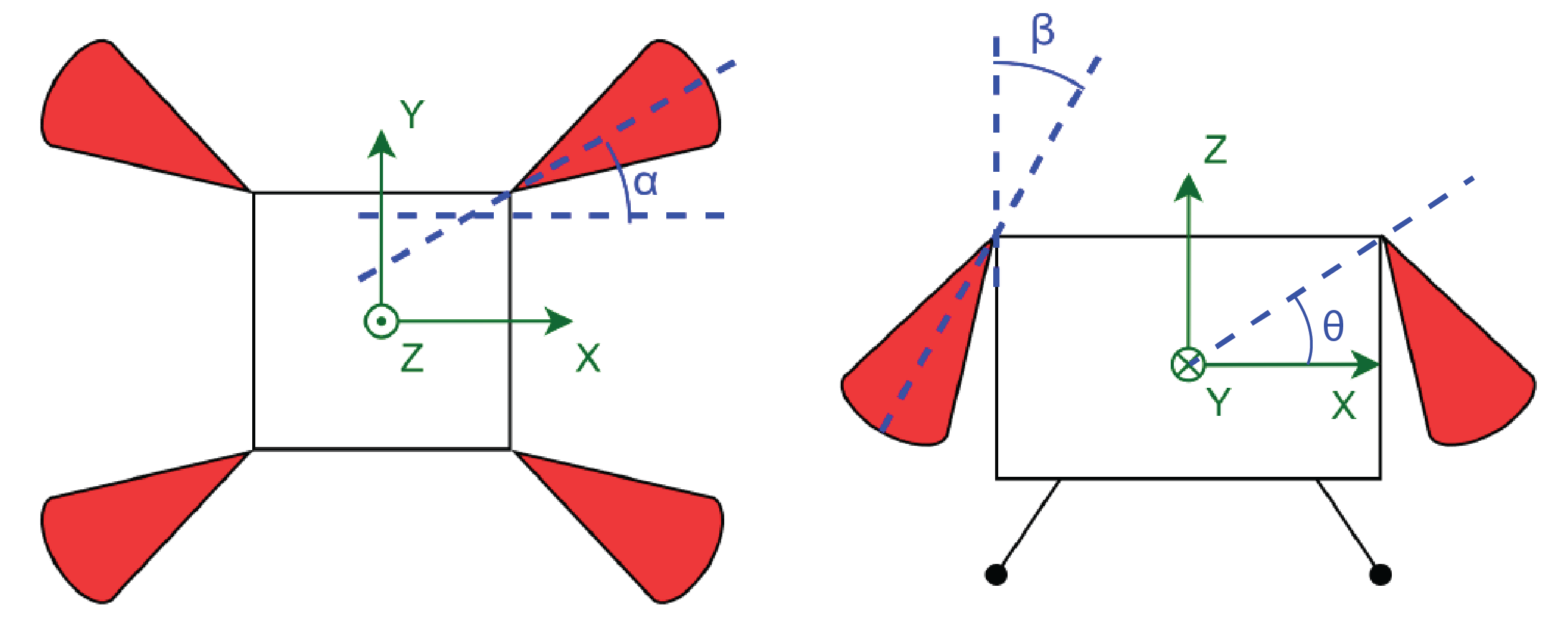}
\end{center}
\caption{Top view (left) and side view (right) of the drone with its four thrusters (in red). X-axis defines the main direction of flight. $\alpha$ and $\beta$ define thruster angles with respect to the x-y and x-z planes, respectively; and $\theta$ defines the drone pitch angle.}\label{fig:drone_config}
\end{figure}

\begin{figure}[h!]
\begin{center}
\includegraphics[width=\textwidth]{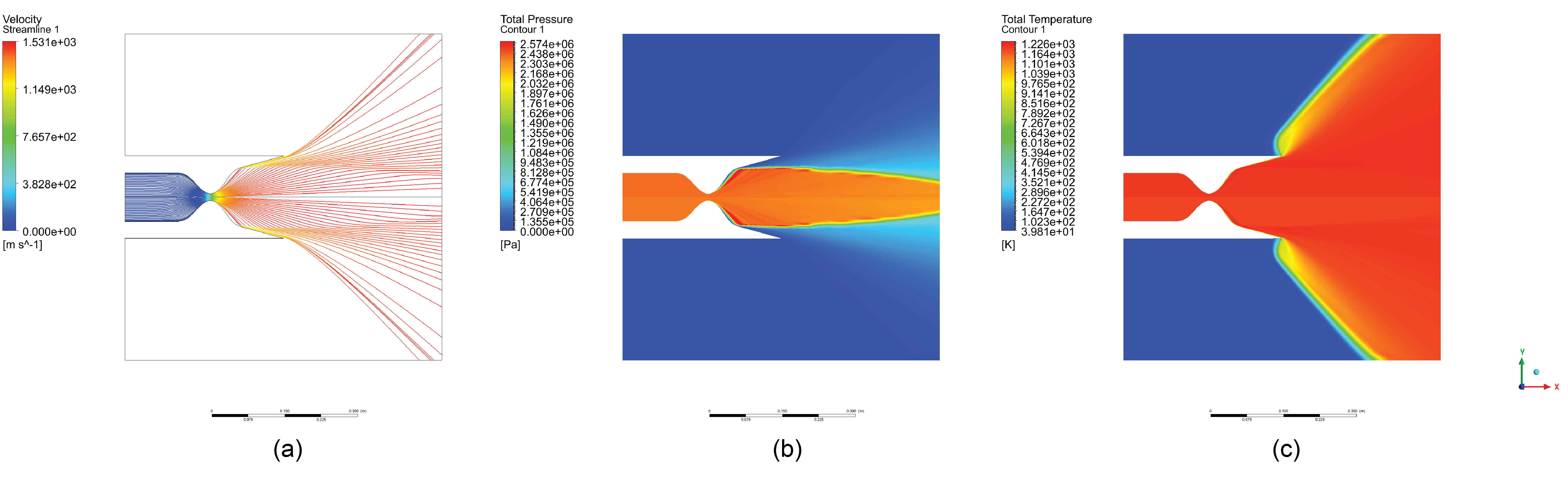}
\end{center}
\caption{Steady-state results of the fluid simulation of one thruster exhaust under vacuum with 2.4\;MPa inlet pressure: (a) streamlines (shape and velocity) of the thruster exhaust gas, (b) total pressure field, and (c) total temperature field.}\label{fig:fluid_sim}
\end{figure}

Unlike the 8-to-1 configuration presented by existing drone concepts, we distributed the engines similar to conventional quadcopter drones with each thruster located on top of the drone, 90-deg from each other. Different placement configurations, at times in combination with reaction wheels (RWs) and control moment gyroscopes (CMGs), were initially simulated and evaluated to find the optimal configuration. Despite the slightly higher propellant consumption of a 4-thruster system, it provides higher controllability at a lower overall mass than single thruster alternatives paired with RWs/CMGs, avoids the need for actuated thrust vector control (TVC) systems \citep{linsen2022}, and allows mounting the mapping sensor at the bottom of the drone pointing nadir. The thrusters are offset 45-deg with respect to the x-axis (angle $\alpha$ in Figure~\ref{fig:drone_config}) to enable precise yawing of the drone and they are angled 45-deg with respect to the z-axis (angle $\beta$ in Figure~\ref{fig:drone_config}). Fuel consumption and total flight time with different thruster angle variations were also evaluated in simulations to find the optimal configuration. While a lower $\beta$ yields lower fuel consumption, the chosen 45-deg configuration provided the most efficient fuel consumption option while minimizing dust dispersion, avoiding disturbances in the measurements, and keeping the outer structure and optics of the drone away from the 38-deg high-temperature and slim high-pressure regions of the engines' exhaust (see Figure~\ref{fig:fluid_sim}). 

The drone makes also use of a regulated helium-based pressurization system to maintain constant pressure in the propellant tank while in flight. Helium is stored in a separate tank and its flow is controlled via a pressure regulator. This system provides higher control over the output pressure and resulting thrust levels compared to blowdown systems, critical for the rapid, precise control of the drone. The drone is equipped with a 1.5-mm titanium spherical bladder tank for the propellant and a 0.6-mm tank of the same material and shape for the pressurant. Tanks are sized for a single 1000-m straight flight based on \cite{larson1991, sutton2017}. Final specifications for the tanks and the pressurization system are listed in Table~\ref{tab:tanks} and all include 20\% margins and a factor of safety (FoS) of 2 to account for potential changes in the overall mass in later iterations, in-flight correction maneuvers, and variations in the trajectory not represented in current simulations (refer to Section~\ref{sec:simulations}). 

\begin{table}[hbt!]
    \centering
    \begin{tabular}{p{6cm} p{2cm} }
        \hline
        \textbf{Specification} & \textbf{Value} \\ 
        \hline
        Propellant consumption per flight & 1.86\;kg \\ 
        Propellant storage pressure & 2.4\;MPa \\ 
        Pressurant storage pressure & 14\;MPa\\ 
        Pressurant total mass & 0.154\;kg\\ 
        Propellant tank mass & 0.91\;kg \\ 
        Propellant tank volume & 3.35\;L \\ 
        Pressurant tank mass & 0.4\;kg  \\ 
        Pressurant tank volume & 1.4\;L \\ 
        Fuel lines length & 3.75\;m \\ 
        Fuel lines mass & 0.288\;kg \\  
        Propellant filter mass & 0.056\;kg \\
        Pressure regulator mass & 0.34\;kg \\
        Pressure/temperature transducers & 0.17\;kg \\
        Fill and drain valves mass & 0.418\;kg \\
        Latch valve mass & 0.1\;kg \\
        Tank and valve heaters mass & 0.168\;kg \\ \hline
    \end{tabular}    
    \caption{Tanks and pressurization system specifications.}
    \label{tab:tanks}
\end{table}

\subsubsection{Mapping instrument}

Five different types of sensors were initially considered: optical camera, radar, scanning LiDAR, flash LiDAR, and thermal infrared camera. We ultimately deemed the use of a flash LiDAR the best option on which to base the design of our lunar drone concept. LiDAR technology achieves higher resolution and performance under rapidly varying lighting conditions than conventional optics and radar. As mentioned before, high-signal, high-resolution images of unlit regions require longer exposure times, a high dynamic range, high frame rates, and, if the aforementioned requirements cannot be met, the use of additional light sources to artificially illuminate the scene. New technologies, such as event-driven cameras \citep{mahlknecht2022} and quantum sensing devices \citep{moritomo2020}, are emerging as promising new technologies with particularly high performances under conditions of poor or rapidly varying lighting and fast movement \citep{rebecq2021, ma2023}. The readiness level for space of these technologies, however, is at the time of writing still low for our baseline design. Unlike conventional scanning or rotating LiDARs, flash LiDARs do not require any moving parts, illuminating the whole scene in single flashes. Currently, lightweight flash LiDARs (\textless\,4\;kg) are being developed for space applications and are expected to become available in the near future \citep{mitev2017, tzeremes2019}. We based our design on the MILA BB model from the Swiss Center for Electronics and Microtechnology (CSEM) \citep{pollini2018} with an objective mass under 2\;kg and maximum power consumption of 35\;W. It is important to note that in order to configure the drone, we assumed the optical elements can be physically separated from the control electronics of the flash LiDAR for better placement and a more compact configuration. 

The selection of the flash LiDAR also introduced the need to fly at a constant altitude of $\sim$\,50\;m (typical LiDAR range) and to do so at a maximum horizontal speed and a maximum pitch angle of 30\;m/s and 24-deg \citep{amzajerdian2011}, respectively. The former was estimated based on the need to comply with R5 (refer to Table~\ref{tab:requirements}) alongside an expected sample rate of 300\;Hz \citep{ball2019, lslidar2022}.

\subsubsection{Electrical power system}

The drone has a maximum peak power consumption of 324\;W. The outcome of a series of flight simulations predicts a total flight time of 140\;s per flight (see Section~\ref{sec:simulations} for details), on top of which a margin of 30\% (i.e., $\sim$180\;s) was used to size the batteries. Imposing 10 battery charge/discharge cycles with a depth of discharge of 90\%, the required battery capacity was estimated to be $\sim$21\;Wh. We ultimately opted for the space-proven iEPS Electrical Power System from ISISpace containing a Lithium-ion battery pack that provides 22.5\;Wh \citep{isispace2019}. It is important to note that the drone was not devised to host an internal power generation system as its batteries will be charged by the service station in between flights (details in Section~\ref{sec:service_station}). 

\subsubsection{Avionics}
 
For the drone avionics, we defined a centralized data architecture in which all the different electronic components are connected to an onboard computer (OBC) (see Figure~\ref{fig:drone_block}). The OBC sends all commands to the active components of the drone and receives housekeeping data from pressure and temperature sensors. The OBC is also in charge of storing all mapping data and the measurements gathered in flight. We used an ISISpace 400 MHz ARM9 OBC as a reference for the design due to its very low weight and power consumption while providing up to 32\;GB of storage \citep{isispace2022}. The drone also makes use of a high-accuracy (biases 0.3$^\circ$/h for the gyroscopes and 0.05\;mg for the accelerometers), low noise (0.15$^\circ/\sqrt{h}$) STIM377H inertial measurement unit (IMU) from SAFRAN to compute the drone attitude during flight \citep{safran2022}.  

\subsubsection{Communications}

The Command \& Data Handling (CDH) subsystem in charge of the communication between the drone and the service station is divided into two modes: in-flight mode and docked mode. The drone is designed to operate fully autonomously. Data is only shared with the service station, which communicates with the serviced rover. Communications with ground stations on Earth, or potentially new lunar orbiting stations, are expected to take place through the rover itself. The data acquired by the flash LiDAR---estimated to be about 20.5\;GB per flight including an expected 25\% compression ratio \citep{kotb2018}--- is temporarily stored by the drone while part of it is directly processed onboard to determine flight parameters, such as position and altitude, and to be used by the hazard detection \& avoidance module during flights in more complex environments such as the inside of lunar pits. The bulk of data acquired in flight will be transferred to the service station by a high-speed data cable after each flight (docked mode). The data sent to the service station during flight is limited to drone health, position tracking, power, and propellant consumption. Basic commands sent through the service station can be also received by the drone during flight such as service station housekeeping and safety checks. For this, the drone makes use of a programmable wireless UHF radio transceiver from Nanoavionics and a Zigbee antenna operating at 440\;MHz and with a maximum bandwidth of 200\;kbps over a line-of-sight up to 1\;km. 

\subsubsection{Thermal control}

\begin{figure}[h!]
\begin{center}
\includegraphics[width=0.8\textwidth]{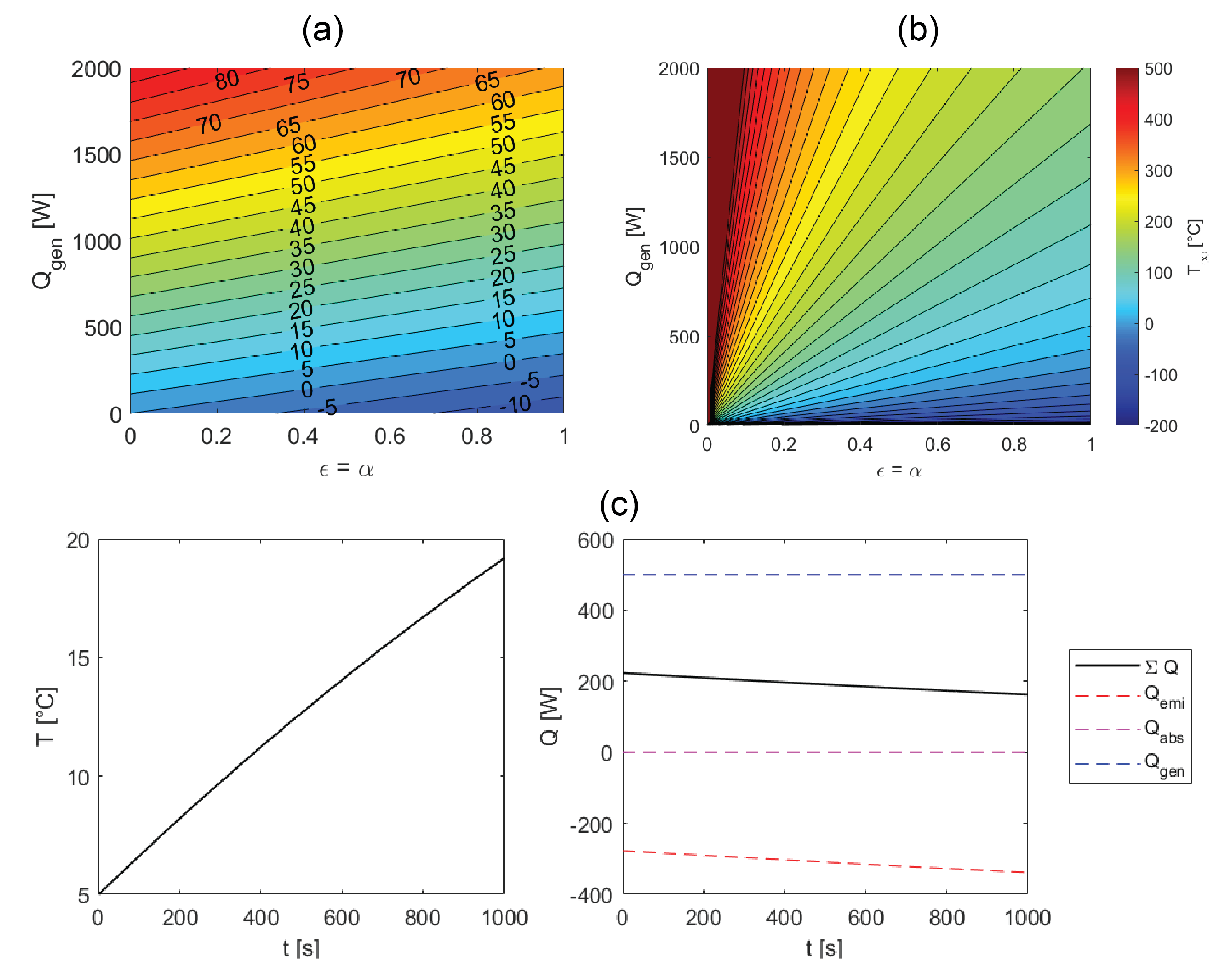}
\end{center}
\caption{Preliminary rate of heat transfer calculation results: (a) drone temperature change for an elapsed time of 600\;s with respect to the thermal source power, $\dot{Q}_{gen}$, and the emissivity/absorptivity; (b) drone equilibrium temperature, $T_\infty$, with respect to $\dot{Q}_{gen}$ and the emissivity/absorptivity; and (c) evolution of the drone temperature, $T$, with respect to time, \textit{t}, for a $\dot{Q}_{gen} = 500$\;W and $\epsilon=\alpha=0.8$. }\label{fig:thermal}
\end{figure}

The thermal design of the drone is particularly challenging, with internal temperatures increasing and dropping rapidly due to the low volume available and the extreme temperatures it may be exposed to during a mission. We conducted a series of preliminary estimations of evacuated thermal power and temperature variations during flight as given by

\begin{equation}
\label{eq:heat_transfer}
    d\mathrm{T} = \frac{\dot{Q}_{rad}+\dot{Q}_{gen}}{m \cdot c_p}\cdot dt,
\end{equation}

where $\dot{Q}_{rad}$ is the radiated heat rate, $\dot{Q}_{gen}$ represents the sum of both incident and internally generated heat, $m$ is the drone mass, and $c_p$ is its heat capacity (assumed to be 900\;J/K). For all the calculations, the background black body radiation is estimated to be emitted at $T_\infty$= 4\;K (deep space), the drone is considered a gray body with emissivity, $\epsilon$, and absorptivity, $\alpha$, equal and constant across all wavelengths, initial uniform temperature of $5^\circ$C, and of a cylinder shape with a surface area of 1.021\;m$^2$. Preliminary calculations of emitted heat transfer rates resulted in a maximum allowed joule heating from electronic components and residual heating from the firing of the thrusters of 500\;W. Beyond this number, heat would not be effectively evacuated without the use of radiators. Figure~\ref{fig:thermal} displays the temperature evolution with respect to time, thermal source power, and emissivity/absorptivity, as well as a detailed temporal evolution of the drone temperature for an estimated $\dot{Q}_{gen} = 500$\;W and $\epsilon=\alpha=0.8$, representative of white paint.

With this in mind, the different components of the drone will be maintained within their operating temperature ranges (see Table~\ref{tab:temperatures}) by means of flexible electrical Polyamide/Kapton heaters. These provide a lower weight, lower power alternative to radiators. The selection of materials needs to be carefully curated to optimize the properties of all passive components (i.e., $c_p$, $\epsilon$, and $\alpha$'s). Multilayer insulation for the external surfaces and thermal straps inside the drone are used to effectively distribute the heat. Special paints and surface coatings can be used to control the emissivity and absorptivity of the different drone surfaces. 

\begin{table}[hbt!]
    \centering
    \begin{tabular}{p{3.5cm} p{4cm} p{5cm} }
        \hline
        \textbf{Subsystem} & \textbf{Component} & \textbf{Temperature Range ($^\circ$C)} \\ 
        \hline
        \multirow{7}{*}{Propulsion} & Fuel/pressurant & [$+8; +53$] \\
        & Pressure regulator & [$-65; +85$] \\
        & Latch valve & [$-25; +50$] \\
        & Fill/drain valve & [$-7.2; +60$]\\
        & Propellant filter & [$-73; +371$]\\
        & Temperature transducer & [$-200; +300$]\\
        & Pressure transducer & [$-40; +60$]\\
        Payload & Flash LiDAR & No data available \\
        Telecommunication & Antenna/Transceiver & [$-40; +85$] \\
        \multirow{2}{*}{Avionics} & OBC & [$-25; +65$] \\
        & IMU & [$-40; +85$] \\
        Power & EPS/Battery pack & [$-20; +70$] \\
        Thermal & Flexible heaters & [$-195; +200$] \\ \hline
    \end{tabular}    
    \caption{Operating temperature range for the selected components.}
    \label{tab:temperatures}
\end{table}

\subsubsection{Structure and wire harness}

The internal structure of the drone is inspired by the design of ESA's Copernicus Sentinel 2a satellite \citep{martimort2007}. It is formed by a skeleton of three composite plates on which all the different elements of the drone are assembled. Carbon fiber legs, similar to those used by NASA's Ingenuity Helicopter \citep{tzanetos2022b}, are fixed to the bottom plate to assist during landing, help position the drone correctly once on the service station (refer to Section~\ref{subsec:tol}), and also touch down safely on the ground in the event of failure requiring an emergency landing away from the service station. The electronics are placed between the tanks of the propulsion system. They contain the optics of the flash LiDAR, the OBC, the EPS, the transceiver, and the IMU, which are all mounted on a single removable electronic stack attached to the vertical plate of the internal structure so that it can be accessed and disassembled easily, facilitating troubleshooting operations during testing. External panels covered in multilayer insulation are used to protect and thermally isolate as much as possible the internal elements of the drone from incident radiation.

\subsection{Drone Service Station}
\label{sec:service_station}

\begin{figure}[h!]
\begin{center}
\includegraphics[width=0.8\textwidth]{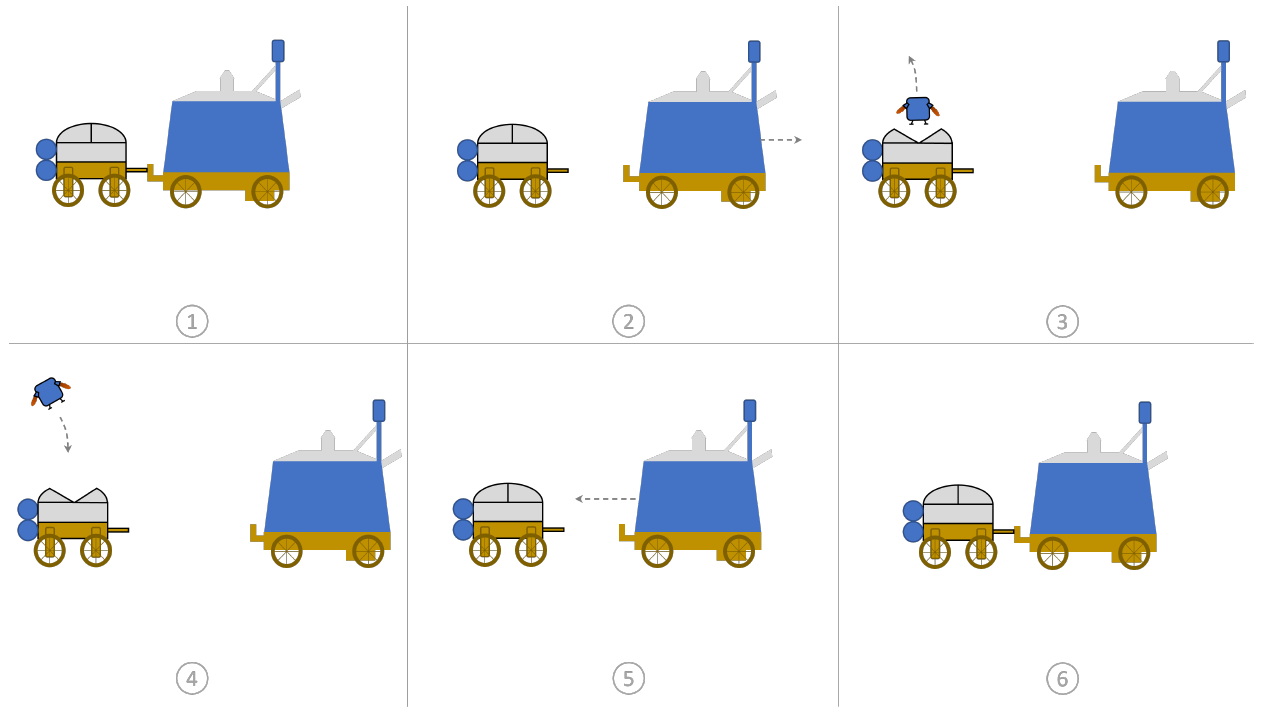}
\end{center}
\caption{Drone service station concept of operations.} \label{fig:dss_conops}
\end{figure}

As previously introduced, the service station was devised as a solution to mitigate most of the mission risks associated with a direct interaction with the lunar surface. While different use cases were initially evaluated (e.g., mounted on top of or deployed by a rover \citep{pfeiffer2022, pozsgay2022}), the final design presents a service station in the form of a 2-wheel towed trailer. This solution increases the overall mass of the drone system and could potentially impair maneuverability but simplifies interfaces, reduces risks and design complexities, and enables the parallel use of the serviced vehicle while the drone is in flight, making operations more efficient (see Figure~\ref{fig:dss_conops}).

\begin{figure}[h!]
\begin{center}
\includegraphics[width=0.8\textwidth]{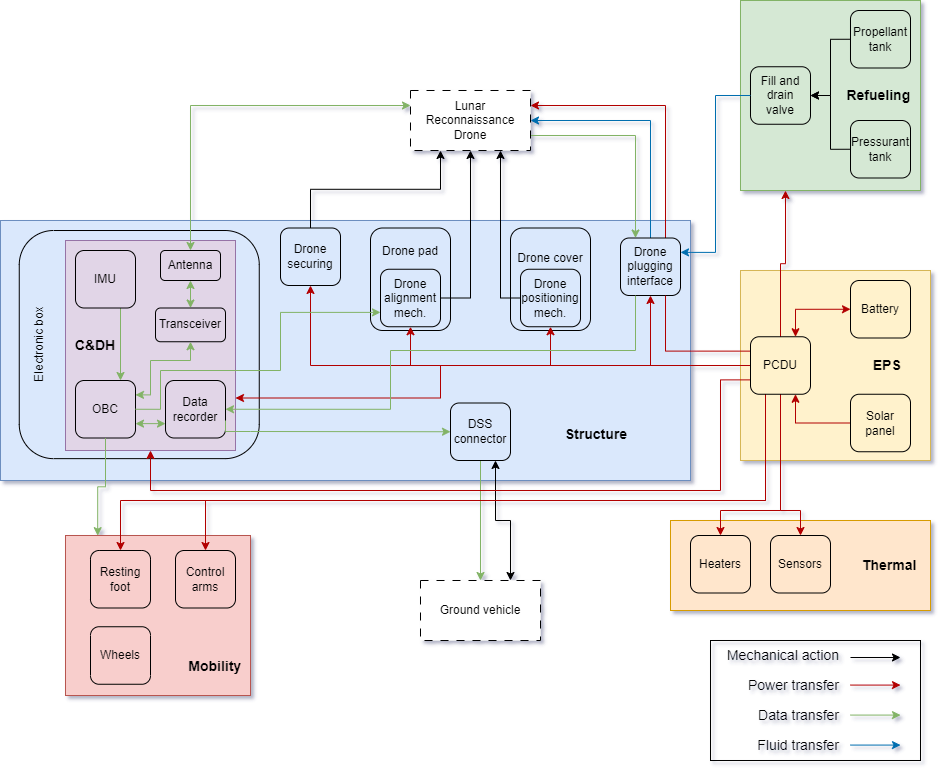}
\end{center}
\caption{Architecture of the drone service station.}\label{fig:dss_arch}
\end{figure}

The drone service station has an overall size of 92 x 92 x 106\;cm and a total wet mass of 83\;kg, of which 32\;kg is devoted to the refueling subsystem (i.e., propellant, pressurant, tanks, fuel lines, sensors, and valves) and 20.5\;kg to the batteries. The service station has been sized to allow the drone to perform ten additional 1000-m flights and to enable the whole system to stay up to 50 hours within shadowed regions in \textit{standby} mode. It is worth highlighting that the system is not designed for the drone to be deployed from within PSRs or other extremely low-temperature regions. The station is powered by high-energy-density batteries providing 246.7\:Wh/kg specific energy \citep{exa2022} fed by 1.29\;m$^2$ of GaInP/GaAs/Ge triple-junction solar cells (considering 29\% efficiency as a reference). The station is also equipped with the same OBC as the drone, paired with a Mercury RH3440 SSD capable of storing 440\;GB of flight data---leaving enough room for all the compressed mapping data acquired over 11 flights and all the housekeeping data from both the drone and the service station---while being flight-proven, compact, and radiation tolerant over 100\;krad \citep{mercury2022}. A simplified version of the station architecture is depicted in Figure~\ref{fig:dss_arch}.

Four elements of the service station are considered key: 1) the refueling subsystem, 2) the towing interface with the serviced vehicle, 3) the TOL pad, and 4) the mobility subsystem of the trailer.

\subsubsection{Refueling}

A safe and automatic connection mechanism between the drone and the storage tanks is needed. Fortunately, as the space industry expands toward more sustainable solutions, so do the technologies devoted to in-orbit servicing (refueling, repair, and maintenance). In our case, the station fill and drain valve design was based on OrbitFab's Rapidly Attachable Fluid Transfer Interface (RAFTI) \citep{orbitfab2022}. This solution can transfer two fluids independently with a maximum misalignment of 4 deg. It is fully compatible with hydrazine and helium at pressures up to 4.48\;MPa and 20.68\;MPa, respectively. Tanks are sized using the same materials as the drone. Apart from the tanks, the refueling subsystem makes use of a pressure regulator, similar to the one used on the drone, a fuel pump, and transducers to control pressure and temperature within the system. Depending on the exact pump selected, refueling can take between 1.5 and 90 minutes. In our case, we opted for a Flight Works 2212-M04C42 M-series pump for its low power consumption, which provides a maximum flow rate of 200\;mL/min. A complete refueling of the drone tanks would, therefore, take slightly over 11 minutes. 

\subsubsection{Towing}

The connection mechanism should be designed so that the service station can attach and detach automatically from the serviced rover as well as to fit a wide variety of ground vehicles and rovers. The towing mechanism could potentially also act as a data interface between the serviced rover and the drone system. After evaluating existing solutions \citep{weiss2020} and due to the lack of flight-proven technologies for this particular use case, we proposed our own design (see Figure~\ref{fig:dss_ss}(b)). The ground vehicle side would feature two vertically actuated, parallel plates with concave cups to house a mating sphere in between, which is attached to the service station. To open the mechanism, the two plates would slide apart via lead screws. The data interface would be fitted in the middle by making the center of the sphere and the protruded beam hollow. The connector restricts pitch rotation to $\pm$25-deg while maintaining free roll rotation and allowing $\pm$80-deg yaw rotation angles to prevent potential issues associated with point turns and tight turns exercised by the towing rover. The spherical mating should compensate for a potential misalignment quite effectively and its diameter can be modified to optimize the design. The effects of dust on the degradation of materials and the performance of the mechanism were not evaluated as part of this study. 

\subsubsection{Take-off \& landing pad}
\label{subsec:tol}
Alongside the towing mechanism, the TOL pad, its surrounding protective plates, and associated opening/closure mechanisms needed to be designed from scratch as no referenced mechanisms could be found in the literature. The service station should not only allow for TOL operations to take place safely but it should also provide a reliable solution for propellant, pressurant, power, and data to be transferred to and from the drone. Correct positioning and alignment of the drone on the pad is, therefore, key. For this, we ultimately opted for a solution that consists of a rotating base mechanism placed on an axial ball bearing and actuated by an electric motor---so as to yaw rotate the drone and align it with the valves and connectors present in the service station---paired with fixed passive pushers with free-rotating heads located at each of the external protective cover plates to translate the drone in the horizontal plane (see Figure~\ref{fig:dss_ss}(c)). This way, minor misalignments in the orientation and position of the drone after landing can be corrected. The main disadvantage of this solution is that the orientation of the drone after landing has to be precisely known. 

An octagonal landing pad is made out of a 4.4-mm 5-layer composite panel. From top to bottom, the landing pad is formed by a 0.15-mm Ti-6Al-4V plate sandwiched between two 0.25-mm high emissivity (top) and low emissivity (bottom) ceramic coating on top of a 2-mm perforated Kapton plate and a Ti-6Al-4V honeycomb sandwiched between perforated plates of the same material. The perforated honeycomb panel configuration is used to minimize the contact surface with the pad itself, thus lowering conductivity. A low emissivity ceramic coating is placed to avoid radiating heat toward the inside of the station and, on the upper side, a ceramic coating with high emissivity is applied to
improve heat resistance. Ceramics can accumulate charges when exposed to radiation, however in this case we consider the drone cover to be closed the vast majority of the time. The landing pad, and the drone, are surrounded and protected when in \textit{standby} by a set of four cover plates, which, when closed, help position the drone in place and protect it from environmental effects and, when open, act as flame diverters to minimize the interaction between the thrusters exhaust and the rest of the station, as well as the generation of dust during take-off and landing. 

\subsubsection{Locomotion}

\begin{figure}[h!]
\begin{center}
\includegraphics[width=0.75\textwidth]{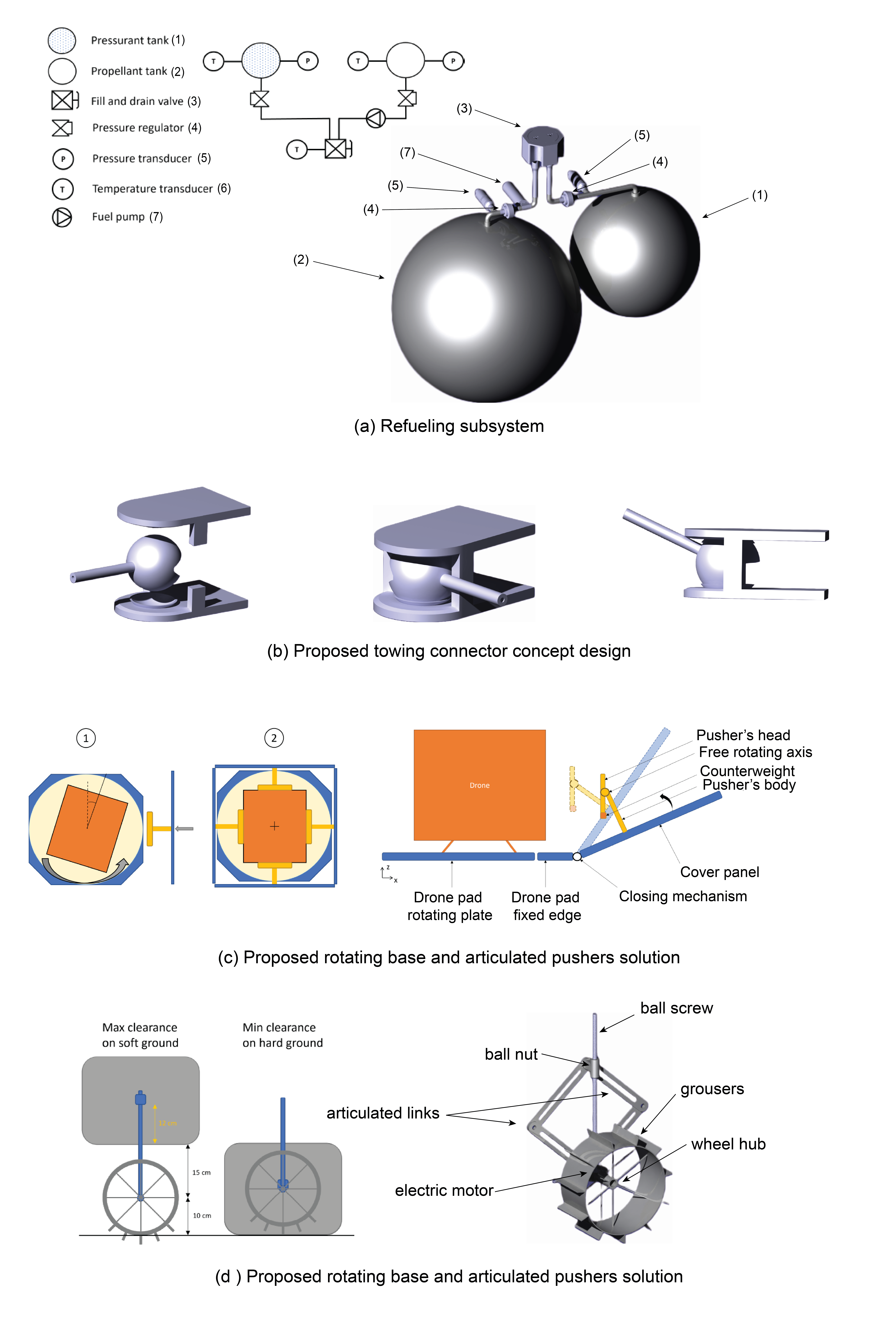}
\end{center}
\caption{Key subsystems of the drone service station.}\label{fig:dss_ss}
\end{figure}

The goal of the locomotion subsystem is to make the service station easily towable by a ground vehicle and robust enough to surmount the irregularities of the lunar surface. The locomotion subsystem, alongside the adjustable resting foot, has been designed so that the ground clearance of the towed station can be adjusted from 0 (stowed configuration) to 30\;cm and the TOL pad can be leveled flat even on slopes up to 20-deg. The adjustable height makes the whole system suitable for transit to the Moon, versatile for easier fitting with varied ground vehicles, and adaptable for effective traversability across uneven terrain profiles. The station makes use of two 20-cm passive Ti-6Al-4V wheels presenting eleven 2-cm Inconel 718 grousers and a resting fore foot. The wheels are each connected to independent control arms actuated by a 36-cm ball screw linear actuator and guided by 2 articulated links in a lozenge shape (see Figure~\ref{fig:dss_ss}(d)). The resting foot is used only when the towing rover unhooks from the trailer. The same ball screw mechanism is used in this case connected to a ground pad. Control arms and resting foot are both covered in flexible Tedlar film to prevent potential damages caused by lunar regolith and dust. 

\section{Flight Trajectory \& Control Simulations}
\label{sec:simulations}

\subsection{Simulation setup}
\begin{figure}[h!]
\begin{center}
\includegraphics[width=\textwidth]{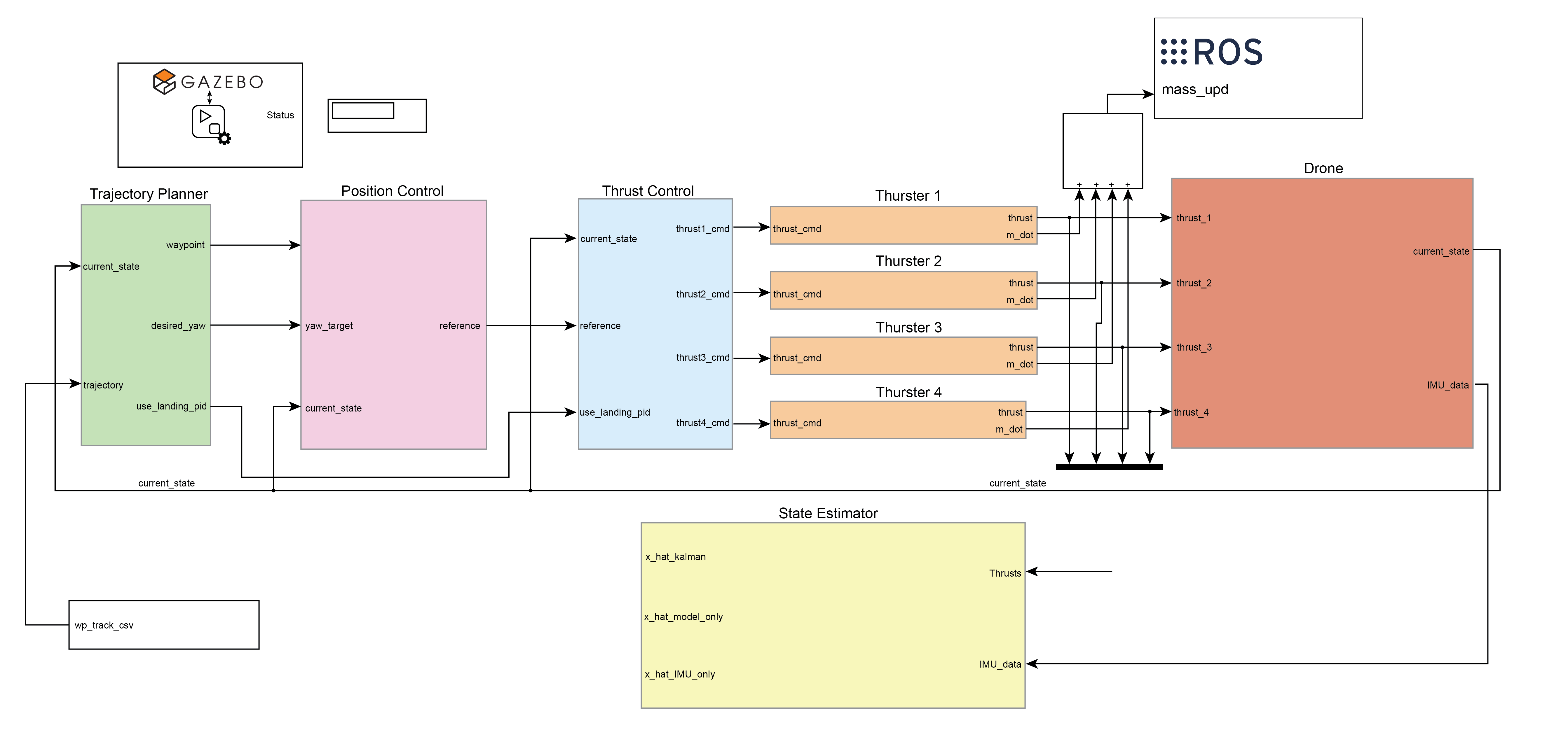}
\end{center}
\caption{Simplified simulation software architecture. }\label{fig:sw_architecture}
\end{figure}

A flight simulation environment was developed using Matlab Simulink to model the drone 2D/3D kinematics and dynamics, Gazebo to visually display the drone and its environment, and ROS to communicate between the two and dynamically modify parameters such as mass flow rate. Six different modules were developed: (1) trajectory planner, (2) position control, (3) thrust control, (4) thruster simulation, (5) drone simulation, and (6) state estimator. A simplified version of the software architecture is shown in Figure~\ref{fig:sw_architecture}. 

The goal of these simulations is to determine flight trajectories, configuration parameters (thruster angles and positions), and drone kinematics/dynamics (position, orientation, velocity, and accelerations) for optimal fuel consumption and flight times. A simplified version of the drone with homogeneous mass distribution and moment of inertia over a perfectly flat ground surface with a global gravity value of 1.62\;m/s$^2$ was used.

\subsection{Flight control and propulsion dynamics}

Monopropellant rocket engines use a liquid fuel contained in a pressurized tank which upon contact with a catalyst produces a high-pressure and high-temperature gas exhausted at very high velocities to generate thrust. The amount of thrust can be computed by

\begin{equation}
    F = \dot{m} \cdot v_e + (P_e - P_a) \cdot A_e, 
    \label{eq:thrust}
\end{equation}

where F is the thrust produced by the engine, $\dot{m}$ is the mass flow rate, $v_e$ is the exit velocity of the exhaust gas, $A_e$ is the exit area of the nozzle, and $P_e$ and $P_a$ are the exit gas pressure and the ambient pressure, respectively. Exit pressure and velocity are defined based on the pressure in the combustion chamber, $P_c$, the exit Mach number, $M_e$ (see Eq.~\ref{eq:mach}), and the specific heat ratio, $\gamma$.
\begin{equation}
    \frac{A_e}{A^*} = (\frac{\gamma + 1}{2})^{-\frac{\gamma +1}{2(\gamma-1)}}\frac{(1+M_e^2 \frac{\gamma - 1}{2})^{\frac{\gamma+1}{2(\gamma-1)}}}{M_e}
    \label{eq:mach}
\end{equation}
\begin{equation}
    \frac{P_e}{P_c} = (1 + M_e^2 \frac{\gamma - 1}{2})^{-\frac{\gamma}{\gamma-1}}
    \label{eq:exit_pressure}
\end{equation}
\begin{equation}
    v_e = M_e \sqrt{\gamma \cdot R\cdot T_e},
    \label{eq:exit_speed}
\end{equation}

where $R$ is the universal gas constant. The mass flow rate is controlled by a valve, simply represented in our simulation by a first-order linear model with a 90\;ms time constant. Note that an important aspect for a precise modeling of the propulsion dynamics is the ignition, which is highly non-linear. Not having access nor the possibility to acquire reliable data to model this transition, we decided to exclude it from the simulations. But associated margins were added to the outcome of these simulations to size the system. The specific design parameters used for the simulations are listed in Table~\ref{tab:prop_sim}.

\begin{table}[hbt!]
    \centering
    \begin{tabular}{p{6cm} p{3cm}}
        \hline
        \textbf{Parameter} & \textbf{Value}\\ 
        \hline
        Drone wet mass & 15\;kg \\
        Combustion chamber temperature & 2800\;K \\
        Combustion chamber pressure & 1.34\;MPa \\
        Ambient pressure & 3e$^{-10}$\;Pa \\
        Specific heat ratio & 1.25 \\
        Throat nozzle diameter & 4.25\;mm \\
        Exit nozzle diameter & 34\;mm \\ 
        Flow rate & 4.1--14.0\;g/s \\
        Gravitational acceleration & 1.62\;m/s$^2$ \\ \hline
    \end{tabular}    
    \caption{Input simulation parameters.}
    \label{tab:prop_sim}
\end{table}

Basic functions of the flight control of the drone (i.e., TOL, stabilization, and waypoint navigation) are achieved via a cascaded Proportional Integral Derivative (PID) architecture with four independent PID controllers to track desired roll, pitch, yaw, and altitude. Position tracking of the drone in fight is implemented via two additional PIDs used to convert instant error in position to a desired roll and pitch angle. No trajectory planning was implemented at this point.

\subsection{Flight trajectories and propellant consumption}

We analyzed propellant consumption on a number of flight profiles, namely: a one-way ballistic hop with a maximum height of 120\;m, a constant-altitude flight with purely vertical and horizontal displacements, and a mixed flight combining a ballistic TOL with a horizontal flight at constant altitude. The baseline for these trajectories is set at 400\;m of total horizontal displacement, a maximum horizontal velocity of 30\;m/s, and a constant flight altitude of 50\;m above ground level (AGL). 


\begin{figure}[h!]
\begin{center}
\includegraphics[width=0.9\textwidth]{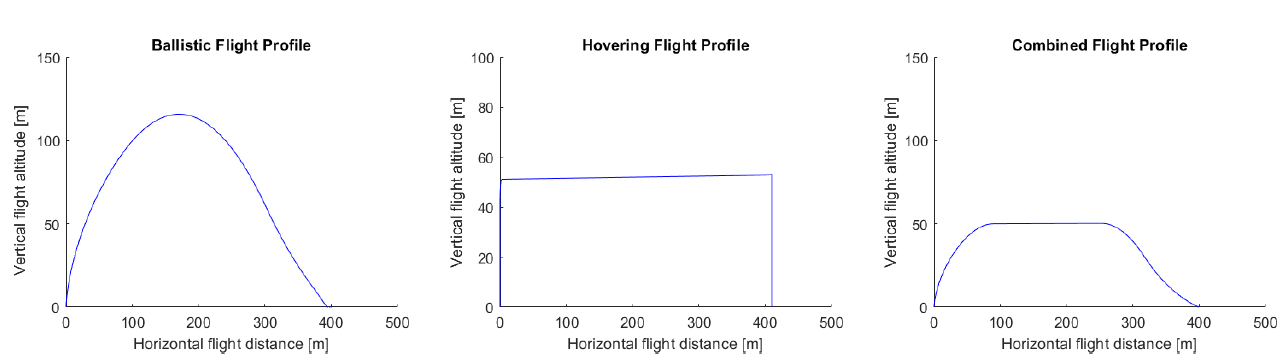}
\end{center}
\caption{Drone trajectories for the three different flight profiles evaluated. }\label{fig:trajectories}
\end{figure}

The results from this preliminary analysis are gathered in Table~\ref{tab:trajectory_results}. Since a constant flight altitude, relatively low flight velocities, and avoiding contact with the ground are of preference for high-resolution mapping of the ground surface, the ballistic trajectory was discarded despite presenting the lowest fuel consumption. The addition of a ballistic TOL to a constant altitude flight profile reduces by 23.8\% total fuel consumption with a more efficient, despite slightly higher, thrust firing. 

\begin{table}[hbt!]
    \centering
    \begin{tabular}{p{2cm} p{2cm} p{2cm} p{2cm} }
        \hline
        \textbf{Flight profile} & \textbf{Average thrust [N]} & \textbf{Total thrust duration [s]} & \textbf{Required propellant mass [kg]} \\ 
        \hline
        Ballistic & 35 & \textbf{6} & \textbf{0.089} \\ 
        Hovering & \textbf{19.125} & 16 & 0.13 \\ 
        Combined & 22.76 & 10.2 & 0.099\\ \hline
    \end{tabular}    
    \caption{Results of preliminary trajectory simulations.}
    \label{tab:trajectory_results}
\end{table}

The resulting optimal trajectory consists of a semi-ballistic take-off and landing---i.e., small vertical TOL with ballistic ascent/descend; a slight alteration of the pure ballistic TOL used in the combined flight profile in Figure~\ref{fig:trajectories}(a)---followed by a constant 50-m altitude flight profile. The short vertical take-off and landing ($\sim$5\;m) provide enough room for correction maneuvers, particularly during landing operations on the service station, which demand high precision and accuracy. Simulations performed with purely ballistic landing experienced an average misalignment of $\sim$0.5\;m with respect to the original take-off location. Semi-ballistic ascents/descents allowed us to reduce fuel consumption by over 13\% compared to a fully vertical TOL. The total mass of propellant consumed during flight (see Figure~\ref{fig:trajectories}(b)) for a total flight distance of 800\;m at a maximum pitch angle of 24-deg and a maximum horizontal speed of 16.68\;m/s is 1.86\;kg for a total flight time of $\sim$140\;s. 

\begin{figure}[h!]
\begin{center}
\includegraphics[width=\textwidth]{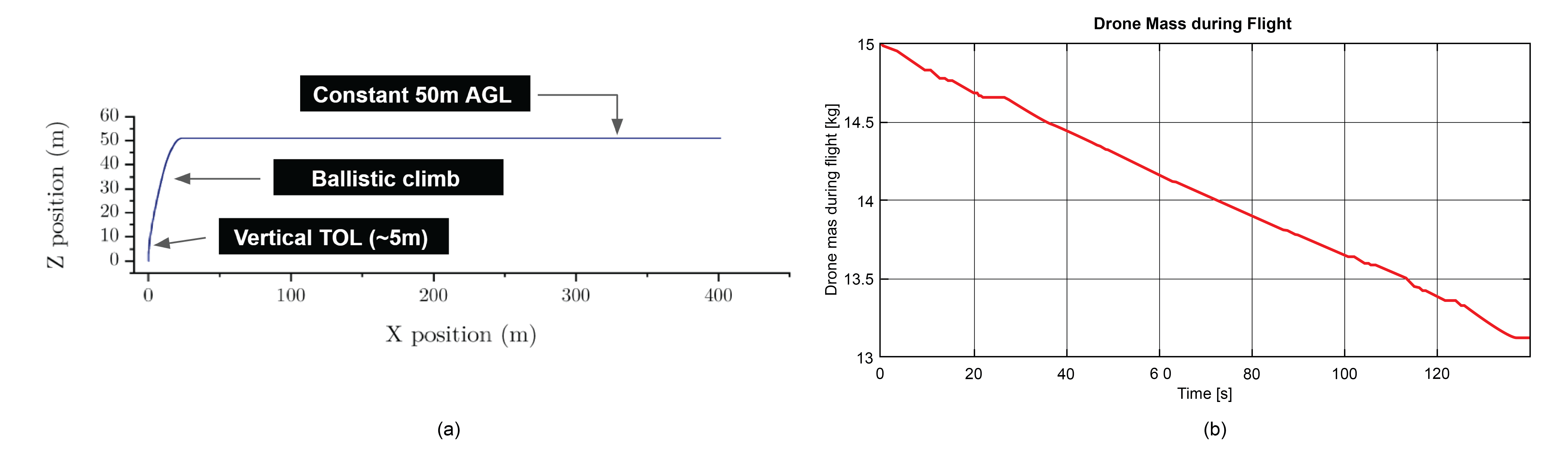}
\end{center}
\caption{(a) Optimized flight trajectory consisting of a semi-ballistic take-off and landing followed (and preceded) by a constant 50-m AGL straight flight profile. (b) Evolution of drone wet mass during flight (an original drone wet mass of 15\;kg was used during simulations).} \label{fig:final_trajectory}
\end{figure}

\section{Evaluation}
\label{sec:evaluation}
Figure~\ref{fig:drone_viper} showcases the potential impact the lunar reconnaissance drone could have if it were to be deployed alongside the upcoming NASA's VIPER mission. The orange arrow indicates the landing zone. Dark green lines defined the planned rover traverse over the 106 days of the mission. The yellow, green, and red areas correspond to different ice depths (surface, shallow, and deep, respectively). Pink arrows were added to the original image to represent single drone flights, with their length at scale for an 800-m round trip flight. We identified three different potential use cases for the drone in this particular scenario: 

\begin{figure}[hbt!]
\begin{center}
\includegraphics[width=0.65\textwidth]{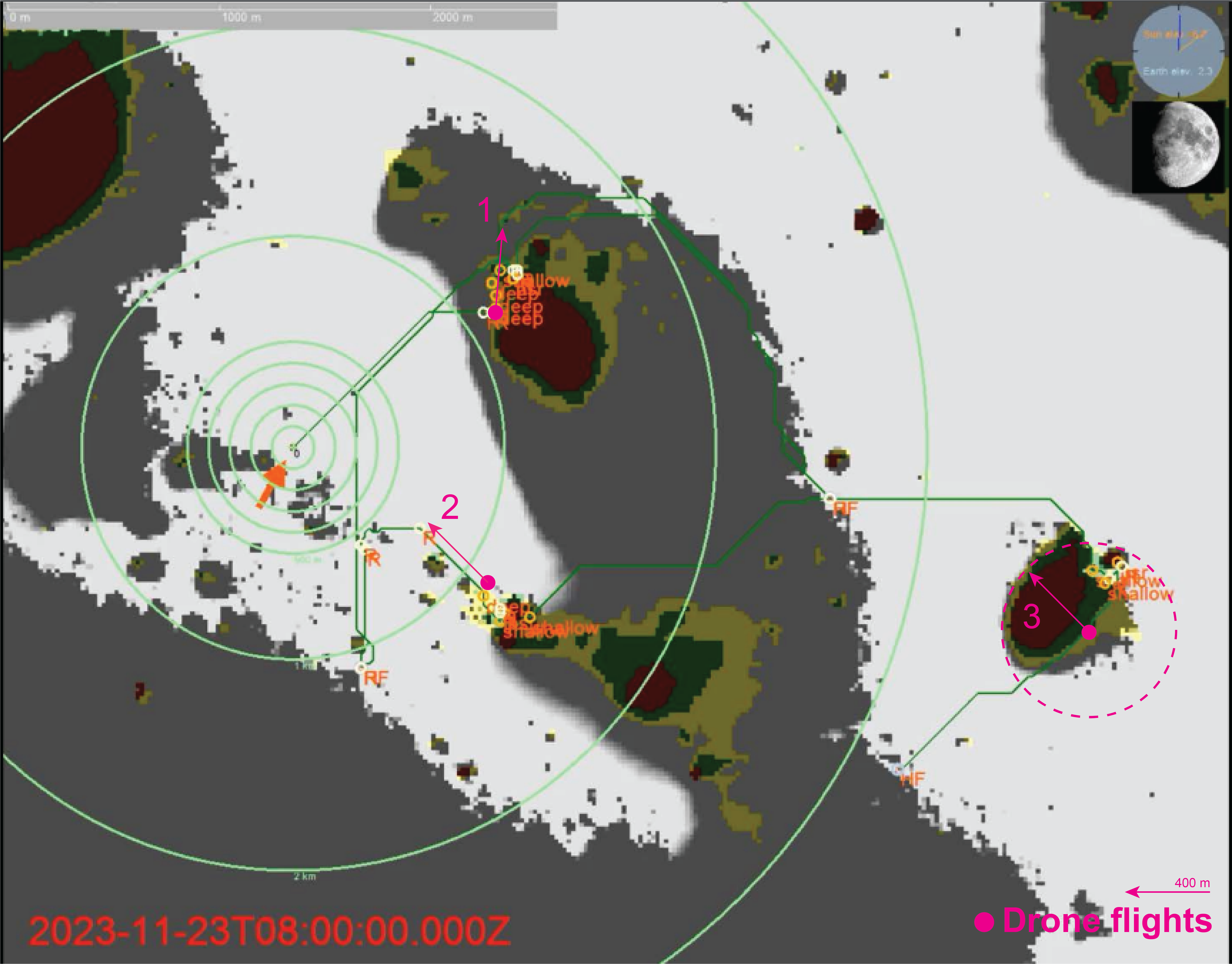}
\end{center}
\caption{Traverse plan of the Nobile region for the upcoming NASA's VIPER mission and potential scenarios for the deployment of the lunar drone (original image: \cite{fong2022}). }\label{fig:drone_viper}
\end{figure}

\begin{itemize}
    \item At location 1, several predefined points of interest are located close to one another. Before the rover stops to examine one of them, the drone can be deployed to fly over the rest, acquire high-resolution images of the surroundings, and perform a preliminary characterization. This would allow the science team to prioritize among the points of interest (order and relevancy), a task impossible to achieve prior to the mission with the data currently available.
    \item At location 2, while the rover stops at its last predefined location, the drone can be already deployed to precisely map the next leg of the traverse. 
    \item At location 3, the drone maximum range is represented as a circular area showing that it is capable of flying over the three types of ice depths in a single 140-s flight. 
\end{itemize}

Given the 4.5--8\;m Waypoint Driving steps of VIPER \citep{colaprete2019}, each requiring data and new commands to be sent and received from the ground to evaluate the path ahead, the deployment of such a system could significantly impact the efficiency of upcoming exploration missions. With the capacity to characterize and map at a high resolution a range of 400\;m per drone deployment, recurrent contact with ground stations on Earth could be drastically reduced to just about one per deployment, i.e., one every $\sim$2\;km of exploration. 

\section{Conclusion}
\label{sec:conclusion}
We described the outcome of a feasibility study and preliminary design of a lunar reconnaissance drone concept aimed at the cooperative exploration of highly relevant and extreme locations on the lunar surface, in particular those of which high resolution (\textless~1\;m/px) geomorphological data does not yet exist. We based the design on upcoming lunar mission requirements, constraints, and priorities. The system consists of a drone and its service station---a 2-wheel towed trailer adaptable to operate alongside different ground vehicles and in charge of providing the drone with a take-off and landing pad, enough propellant, pressurant, and power for additional flights, and shelter from extreme temperatures and radiation when not in operation. We described in depth the design of the drone and provided high-level specifications of the whole system while sharing low-level details of key subsystems of the service station. 

The results presented showcase the feasibility of the design and its expected impact. With under 100\;kg of total wet mass (inc. the service station), the drone system is capable of performing 11 flights, mapping a total horizontal distance of $\sim$9\;km without refueling the station. Space-proven, high-TRL, off-the-shelf components were used as a reference whenever possible. The custom design of certain elements was kept to a minimum and only used when no flight-proven solution could be found in the literature. 

Given the preliminary nature of these results, some limitations are worth highlighting: 1) \textbf{reusability} is one of the core principles of the presented concept. While hydrazine-based thrusters were chosen as a baseline for the design of the drone due to its high TRL and commercial availability, it is our intention for the design to evolve toward more sustainable and reusable engines (e.g., $\mathrm{H_2O_2}$-based propellants \citep{davis2014}); 2) \textbf{simplified simulations and analyses} were performed to achieve rough-order estimations of certain sizing values (mass, volume, power, data). In particular, a more exhaustive thermal characterization of the system would be required in upcoming phases of the project; 3) an extensive \textbf{modeling of the performance of the engines} was conducted but data is still required to define and develop specific control approaches and assess potential failure modes. Additional information is necessary with respect to the engines' ignition, throttleability, and degradation of the catalyst over time, information that at times is only available through testing. 

The Lunar Reconnaissance Drone concept and the results presented herein showcase the need for innovative solutions that can significantly impact the efficiency of upcoming exploration missions by providing already planned and future missions with sub-meter resolution maps of relevant regions of interest.



\bibliographystyle{elsarticle-num} 
\bibliography{biblio}






\end{document}